\documentclass{article}
\usepackage{spconf,amsmath,graphicx,hyperref}

\usepackage{algorithmic}
\usepackage{graphicx}
\usepackage{textcomp}
\usepackage{xcolor}
\usepackage{booktabs}
\usepackage{adjustbox}
\usepackage{graphicx}
\usepackage{mathtools}
\usepackage{subcaption}
\usepackage{xcolor}
\usepackage{multirow}
\usepackage{changepage}

\usepackage{cite}
\usepackage{amsmath,amssymb,amsfonts}
\usepackage{graphicx}
\usepackage{booktabs}
\usepackage{multirow}
\usepackage{adjustbox}

\title{Zero-shot video highlight detection based on text descriptions  and synthetic images}

\name{Michal Byra$^{1,2,*}$\thanks{*Corresponding author: m.byra@samsung.com} \quad Alberto Presta$^{1}$ \quad Grzegorz Stefanski$^{1}$  \quad Krzysztof Arendt$^{1}$ }
\address{$^{1}$Samsung AI Center, Warsaw, Poland  \quad \\ $^{2}$Institute of Fundamental Technological Research, Polish Academy of Sciences, Warsaw, Poland}

\begin{document}
%
\maketitle
\begin{abstract}
Detecting video highlights, the most informative or engaging moments in a video, is important for applications such as video summarization and content recommendation. We propose a zero-shot framework that combines CLIP, large language models (LLMs), and diffusion models. Given lightweight video metadata, such as a title or category, an LLM generates textual descriptions of likely highlight events. These descriptions are further converted into synthetic visual prototypes using a diffusion model. Textual and visual representations are matched to video frames using CLIP, enabling frame-level highlight detection without highlight annotations or dataset-specific training. Experiments on TVSum and SumMe demonstrate strong zero-shot performance, with particularly favorable results on TVSum. The proposed approach provides an effective framework for metadata-conditioned zero-shot video highlight detection.
\end{abstract}

\section{Introduction}
\label{sec:intro}

Video highlight detection aims to identify the most engaging or informative segments of a video and supports applications such as summarization, recommendation, editing, and video search. Most existing methods rely on supervised learning with human-annotated summaries \cite{song2015tvsum,zhang2016video,cai2018weakly}, while unsupervised approaches exploit criteria such as clustering, representativeness, or diversity \cite{tang2023deep,wang2020learning}. More recently, zero-shot summarization has been formulated using contrastive criteria computed from pretrained visual features \cite{pang2023contrastive}. However, such methods usually analyze relationships within the video and lack external semantic knowledge about events likely constituting meaningful highlights.

Multimodal approaches provide additional semantic context. Highlight-CLIP exploits pretrained CLIP representations \cite{Han2024,radford2021learning}, while cross-modal transformers \cite{Xu2024}, audio-visual models \cite{ye2021temporal}, diffusion-based retrieval \cite{zhao2024diffusionvmr}, and multimodal LLMs \cite{lee2025video} have also been explored for video understanding tasks. Hu et al. \cite{Hu_2025_CVPR} use an M-LLM-based selector to identify query-relevant frames for video question answering rather than query-free human-perceived highlights. These approaches generally require task-specific training, video-level queries, or analysis of video content during highlight scoring.

In this work, we propose a zero-shot framework that predicts video highlights from lightweight metadata, such as a title or category, using pretrained foundation models. An LLM generates text descriptors of likely highlight events, which are matched to video frames using CLIP to estimate importance scores. We further translate the descriptors into synthetic visual prototypes using a diffusion model and compare them with video frames in CLIP image-embedding space. The diffusion model can additionally be conditioned on a high-scoring frame to improve visual alignment with the target video.

The proposed framework therefore imitates how a person may imagine likely highlights after receiving only a brief description of a video. It requires neither video-specific training nor highlight annotations, while combining semantic text descriptions with synthetic visual examples. Experiments on TVSum and SumMe demonstrate strong zero-shot performance: LLM-generated descriptors substantially improve over direct metadata matching, while synthetic images provide a complementary signal for highlight detection.

\begin{figure*}[t]
    \centering
    \includegraphics[width=0.8\textwidth]{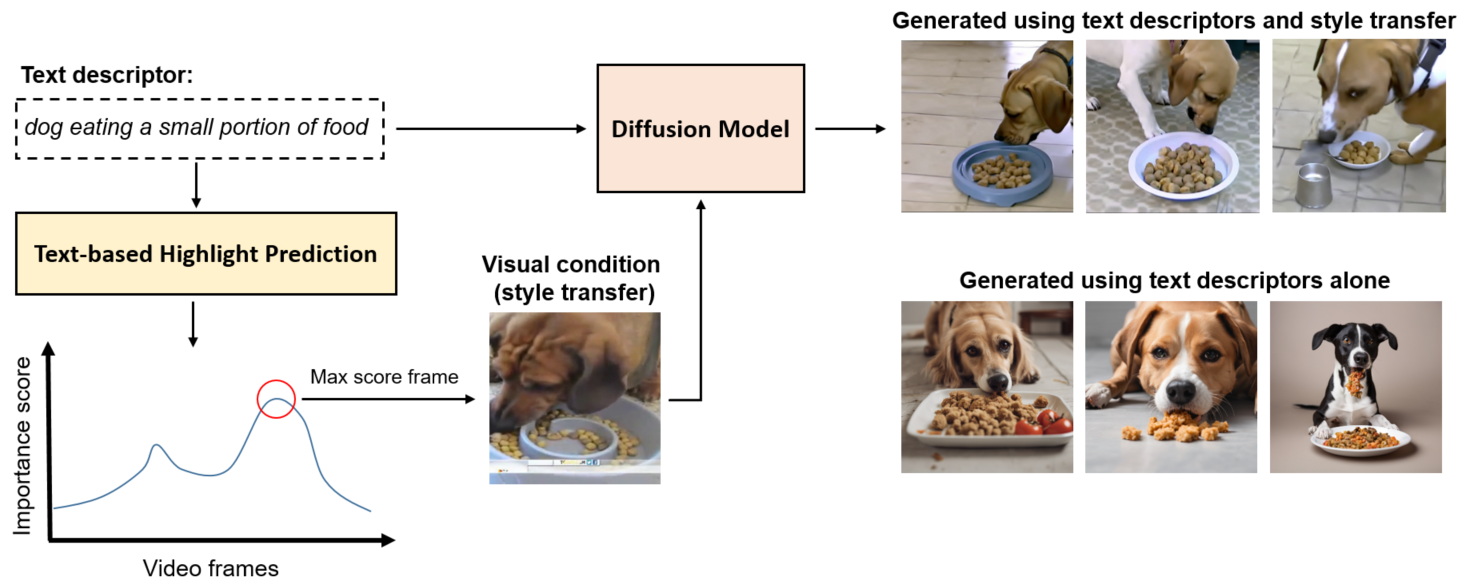}
    \caption{Zero-shot highlight detection using LLM-generated text descriptors and diffusion-generated visual descriptors. The highest-scoring frame under the text-based score can condition image generation, aligning the synthetic descriptors with the appearance of the target video.}
    \label{fig:diffusion}
\end{figure*}

\section{Method}
\label{sec:method}

Given video metadata, such as its title and category, our method uses an LLM to generate $D$ text descriptors $\mathbf{d}=\{d_1,\ldots,d_D\}$ describing likely highlights. These descriptors are compared with individual video frames using CLIP and are also translated into synthetic visual descriptors using a diffusion model. The resulting text- and image-based similarities are combined into a frame-level highlight score. The complete framework is illustrated in Fig.~\ref{fig:diffusion}.

\subsection{Text and Visual Highlight Scores}

First, the LLM is prompted to output visually grounded descriptions that correspond to the best video moments:

\emph{Given the following video metadata, list short and detailed visual descriptions of frames corresponding to the best moments. The descriptions should contain features recognizable by a CLIP-like model...}

Let $x\in\mathbb{R}^{W\times H\times c}$ denote a video frame. Its text-based highlight score is computed as follows:

\begin{equation}
    s_T(x)=\frac{1}{D}\sum_{i=1}^{D} f_T(d_i,x),
    \label{eq:text_score}
\end{equation}

\noindent where $f_T$ is the CLIP similarity computed using its text and image encoders. Frames strongly aligned with the generated descriptions receive high importance scores.

\begin{figure}[b!]
  \centering
  \includegraphics[width=2.8in]{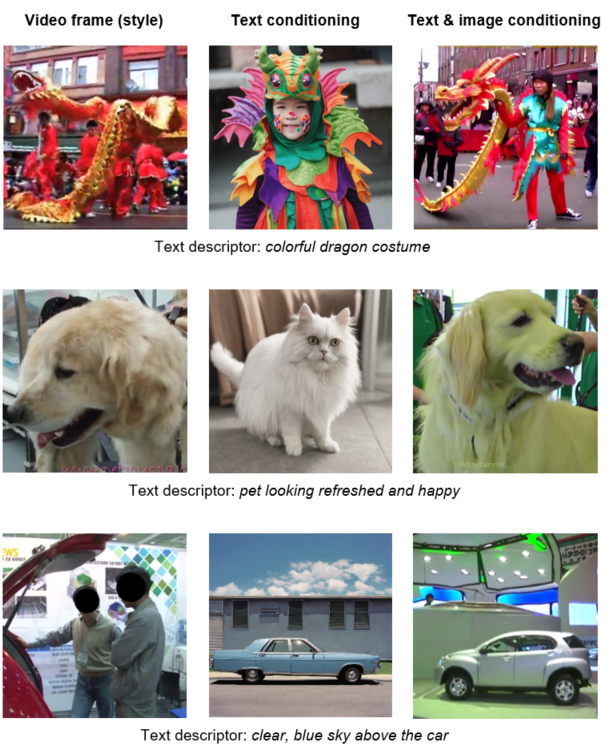}
  \caption{Video frame conditioning may be used to generate synthetic images that improve video highlight detection. Diffusion models may produce images that are not fully aligned with the actual contents of the video. Here, image conditioning brings extra context for the search for video highlights.}
  \label{fig:example}
\end{figure}

We additionally translate each descriptor $d_i$ into a synthetic image using a diffusion model $H$. The generation can be conditioned only on the text descriptor or additionally on a reference frame $x_s$ from the target video. The corresponding image-based score is as follows:

\begin{equation}
    s_I(x,x_s)=\frac{1}{D}\sum_{i=1}^{D}
    f_I\!\left(H(d_i,x_s),x\right),
    \label{eq:image_score}
\end{equation}

\noindent where $f_I$ is the similarity score between CLIP image embeddings. Text only image generation is done by omitting $x_s$.

To obtain a video-specific reference, we select the frame with the highest text-based score:

\begin{equation}
    x_s=\arg\max_x s_T(x).
\end{equation}

Conditioning the diffusion model on this frame transfers video-specific appearance, objects, and texture to the generated images, potentially improving their alignment with the target content, see Fig. 2 for examples of synthetic images. 

Next, text and visual scores are combined descriptor-wise:

\begin{equation}
    s(x)=\frac{1}{D}\sum_{i=1}^{D}
    f_T(d_i,x)\,
    f_I\!\left(H(d_i,x_s),x\right).
    \label{eq:combined_score}
\end{equation}

\noindent The multiplicative formulation promotes frames that are simultaneously consistent with the semantic description and its synthetic visual realization. Frames are ranked according to $s(x)$ to obtain the predicted highlights.

\subsection{Evaluation Protocol}
\begin{table}[b!]
\caption{Correlation scores for the proposed methods. }
\centering
\resizebox{\columnwidth}{!}{
\begin{tabular}{lcccc}
\toprule
 & \multicolumn{2}{c}{TVSum} & \multicolumn{2}{c}{SumMe} \\
\cmidrule(lr){2-3} \cmidrule(lr){4-5} 
& Kendall & Spearman & Kendall & Spearman \\

\midrule

Video title             &  0.141          &    0.184        & 0.052            & 0.064  \\
Video category          & 0.161 & 0.211       & 0.003          & 0.004 \\
Text descriptors        & 0.203 & 0.266            & \textbf{0.088}                 & \textbf{0.108}        \\
Image descriptors                    & 0.205 & 0.267   & 0.056            & 0.069   \\
Text \& image descriptors                  & 0.213 & 0.278   & 0.081           & 0.099   \\
Text \& image descriptors, style transfer         & \textbf{0.217} & \textbf{0.283}   & 0.080            & 0.098   \\

\midrule
\end{tabular}
}
\label{t1}
\end{table}

\begin{table*}[h!]
\caption{Comparison between the zero-shot approach proposed in this work and the previous methods.}
\centering
\scalebox{0.75}{
\begin{tabular}{clcccc}
\toprule
& & \multicolumn{2}{c}{TVSum} & \multicolumn{2}{c}{SumMe} \\
\cmidrule(lr){3-4} \cmidrule(lr){5-6} 
& & Kendall & Spearman & Kendall & Spearman \\
\midrule
& Human baseline          & 0.176           & 0.202           & 0.180  & 0.186 \\
\midrule
\multirow{3}{*}{ Supervised }
& VASNet  \cite{fajtl2018summarizing}  & 0.169           & 0.222           & 0.022          & 0.026  \\
& dppLSTM  \cite{zhang2016video}        & 0.030           & 0.038           & -0.026          & -0.031 \\
& Multi-ranker \cite{saquil2021multiple}                   & 0.176  & 0.230 & 0.011          & 0.014  \\
\midrule
\multirow{6}{*}{ Unsupervised } 
& $\text{DR-DSN}_{60}$ \cite{zhou2018deep}    & 0.017           & 0.023           & 0.043           & 0.050  \\
& $\text{DR-DSN}_{2000}$    & 0.152           & 0.198            & -0.016          & -0.022 \\
& $\text{SUM-FCN}_{unsup}$ \cite{rochan2018video}  & 0.011           & 0.014           &0.008            & 0.010  \\
& SUM-GAN  \cite{apostolidis2020ac}               
                                                             & -0.054          & -0.070          & -0.009          & -0.012 \\

& Contrastive learning (alignment)  \cite{pang2023contrastive}                                                  
                                                             & 0.100           & 0.132           & 0.094           & 0.115 \\

& Contrastive learning (alignment, uniformity, uniqueness)  
                              & 0.161      & 0.212          & 0.036           & 0.044 \\
\midrule

\multirow{3}{*}{ Zero-shot} 
& Contrastive learning (alignment)                   & 0.106    & 0.139  & 0.096  & 0.117 \\

& Contrastive learning (alignment, uniformity)   
                                                             & 0.135           & 0.178           & 0.082         & 0.100 \\

& Proposed (best variant)     
                                                             &  {0.217}          &   {0.283}     & 0.088        & 0.108 \\

\bottomrule
\end{tabular}
}
\label{t2}
\end{table*}

We evaluate the method on TVSum and SumMe~\cite{song2015tvsum,gygli2014creating}. TVSum contains 50 videos from 10 categories, with 20 annotators assigning importance scores from 1 to 5 to two-second clips. Since annotators were shown each video's title and category, TVSum directly matches our native metadata-conditioned setting. SumMe contains 25 videos with 15--18 human summaries, each covering approximately 5--15\% of the video, but does not provide video categories. We therefore evaluate SumMe in an auxiliary bootstrapped-metadata setting, in which Qwen2.5-VL assigns each video a short generic category, such as ``animals'' or ``sports''~\cite{bai2025qwen2}. The inferred category is used only to generate highlight descriptors and is not involved in frame-level scoring. Therefore, TVSum represents the native metadata-conditioned setting, whereas SumMe evaluates whether coarse metadata can be automatically bootstrapped when it is unavailable.

Following previous work~\cite{otani2019rethinking,pang2023contrastive}, we primarily report Kendall's $\tau$ and Spearman's $\rho$ between predicted and human importance scores. Correlations are computed separately for each annotator and then averaged across annotators and videos. We additionally report top-5 mAP on TVSum for comparison with earlier studies. No highlight annotations or dataset-specific training are used for either dataset.

Similarity scores are computed using CLIP ViT-L/14~\cite{radford2021learning}. GPT-o4 generates an average of 12.8 descriptors per video, while synthetic images are produced using SDXL with IP-Adapter~\cite{podellsdxl,ye2023ip}. The image-conditioning scale is set to $0.7$ by default and varied from $0$ to $1$ in the ablation study.

\section{Experiments}

\begin{table*}[!h]
\centering
  \caption{Top-5 mAP scores obtained for the investigated methods on the TVSum dataset. Each column represents a specific video category from the dataset, i.e. VT stands for \emph{changing Vehicle Tire}.}
  \scalebox{0.75}{
  \begin{tabular}{c | cccccccccc | c}
    \textbf{Method} & VT & VU & GA & MS & PK & PR & FM & BK & BT & DS & Mean \\ \hline  
    Human baseline & 0.603 & 0.548 & 0.597  & 0.532  & 0.500  & 0.524  & 0.478  &   0.511& 0.602  & 0.479  & 0.537 \\ \hline
    Video title & 0.437 & 0.514 & 0.166  & 0.609  & 0.426  & 0.441  & 0.583  &   0.381 &  0.431 & 0.417  & 0.440  \\
    Video category  & 0.630 & 0.366 & 0.676  &  0.489 & 0.436  & 0.398  & 0.559  &  0.526 & 0.570  & 0.220  &  0.487 \\
    Text descriptors & 0.587 & 0.578 & 0.739  & 0.516  & 0.519  & 0.622  & 0.358  & 0.644  & 0.502  &  0.467  & 0.553  \\
    Synthetic images & 0.674 & 0.525 & 0.644  & 0.583  & 0.474  & 0.526  & 0.287   &  0.628 &  0.739 & 0.485  & 0.557   \\    
   Text descriptors \& synthetic images & 0.676 & 0.509 & 0.671  & 0.620  & 0.555  &  0.588 & 0.371  & 0.602  &  0.678 & 0.458  & 0.573  \\
   Image descriptors, style transfer & 0.651 & 0.491 & 0.696  &  0.606 &  0.502 &  0.567 & 0.344  & 0.555  & 0.651  & 0.481  & 0.555  \\  
   Text and image descriptors, style transfer & 0.659 & 0.518 &  0.642 & 0.544  & 0.536  &0.526   & 0.392   & 0.602  & 0.660  &  0.369 &  0.545 \\
   \midrule
\end{tabular}
}
\label{t3}
\end{table*}

\subsection{Highlight Score Modeling}

Results in Table \ref{t1} show a clear performance progression on TVSum. Direct CLIP matching using only the video title or category provides a relatively weak baseline. Expanding the metadata into LLM-generated text descriptors substantially improves the correlation scores, indicating that the LLM can infer actions and events likely to correspond to highlights. Synthetic visual descriptors generated from these texts achieve comparable performance, showing that imagined visual prototypes can also provide useful cues. Combining textual and visual descriptors yields further gains, suggesting that the two modalities provide complementary information. The best correlation results are obtained when both descriptor types are used together with style transfer, reaching Kendall and Spearman scores of 0.217 and 0.283, respectively.

However, the correlation scores obtained on SumMe were lower than on TVSum. SumMe does not provide native category metadata, so the descriptors are generated from automatically inferred coarse categories rather than dataset-provided metadata. The best performance on SumMe was obtained using the LLM-generated text descriptors. Moreover, SumMe titles are often too short and generic to provide useful visual cues, e.g., “Air Force One” or “Jumps.” Similar to the observations reported in a previous study \cite{pang2023contrastive}, the lower correlation scores may also be related to the annotation procedure used to develop SumMe, where human raters were asked to select between 5\% and 15\% of the video as highlights, resulting in relatively sparse annotations. This sparsity makes it more difficult to compute meaningful correlation coefficients, as the majority of frames are labeled as non-important.

Table \ref{t2} compares the performance of our best approach with that of selected methods from previous studies, following  \cite{pang2023contrastive}. On the TVSum dataset, our zero-shot technique outperformed all previous methods, including both supervised and unsupervised approaches. The proposed method also exceeded the leave-one-out human agreement baseline under the correlation metrics, indicating strong alignment with human highlight annotations. On the SumMe dataset, while our method did not surpass the human baseline, it achieved better results than most previous methods, and performed on par with the zero-shot method proposed by \cite{pang2023contrastive}.

\begin{table}[!b]
\caption{Diffusion strength ablation.}
\centering
\scalebox{0.80}{
\begin{tabular}{ccccc}
\toprule
& \multicolumn{2}{c}{TVSum} & \multicolumn{2}{c}{SumMe} \\
\cmidrule(lr){2-3} \cmidrule(lr){4-5} 
Conditioning strength & Kendall & Spearman & Kendall & Spearman \\

\midrule

0        &       0.205     &    0.267       & 0.056          & 0.069  \\

0.3          &  0.212        &   0.276        & 0.057            & 0.070  \\
0.5          & 0.213 & 0.278       & 0.052          & 0.063 \\
0.7        & 0.214 & 0.278            & 0.050                 & 0.062        \\
1                   & 0.214 & 0.278 & 0.047            & 0.057   \\

\midrule
\end{tabular}
}
\label{t4}
\label{tvsum_spearman}
\end{table}

\begin{figure}[b!]
  \centering
  \includegraphics[width=2.8in]{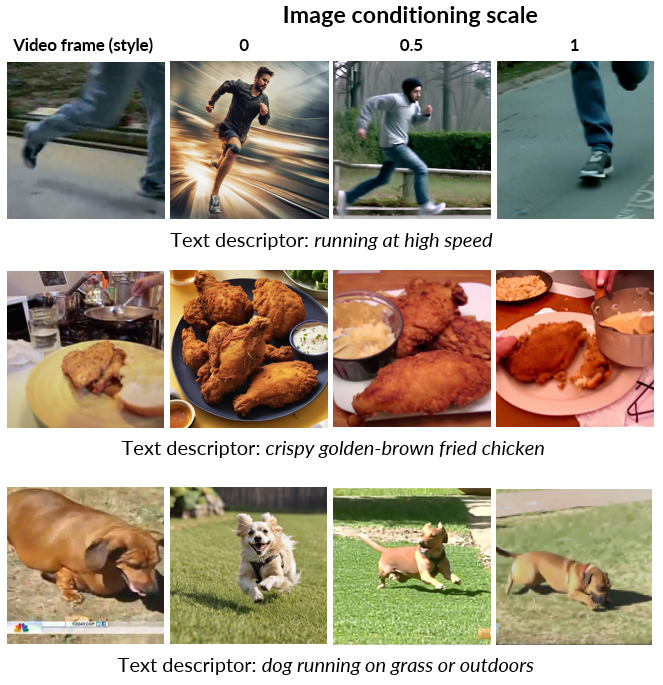}
  \caption{Effect of the image-conditioning strength on diffusion-generated synthetic images. Increasing the conditioning strength more strongly transfers visual characteristics of the selected video frame.  }
  \label{fig:ablation}
\end{figure}

\subsection{Highlight Localization}

To complement the correlation scores, we also computed the mean average precision (mAP) metric on TVSum. Results obtained for the investigated methods are presented in Table~\ref{t3}. Human baseline scores were determined using leave-one-out cross-validation. Here, we can observe that the proposed methods achieved performance on par or slightly better compared to the  human baseline. While a high mAP value may be influenced by the segment extraction procedure and could be large even for random scores, our results further emphasize the usefulness of the proposed approach.

\subsection{Ablation Study}

Results of the diffusion model ablation are presented in Table~\ref{t4}. Increasing the image-conditioning strength improves correlation scores on TVSum, while no improvement is observed on SumMe. This suggests that video-specific conditioning can improve the usefulness of synthetic visual descriptors, depending on the dataset. The effect on generated images is illustrated in Fig. \ref{fig:ablation}.

\section{Conclusions}

n this work, we introduced a novel zero-shot framework for video highlight detection that leverages video metadata and foundation models to estimate frame-wise importance scores. On TVSum, our approach achieved \textit{state-of-the-art} correlation performance, outperforming previous supervised and unsupervised methods. To the best of our knowledge, this is the first study to use synthetic images as imagined visualizations of potential video highlights.

While the method achieved strong results, it relies on metadata such as video titles and categories, which may not always be available. Future work may therefore incorporate video tagging or scene/object recognition to infer coarse video content automatically. The framework can also accommodate user-specific preferences through prompting and benefit from more specialized foundation models.

{\small
\bibliographystyle{IEEEbib}
\bibliography{strings,refs}
}

\end{document}